\documentclass[10pt,twocolumn,letterpaper]{article}

\usepackage{cvpr}
\usepackage{times}
\usepackage{epsfig}
\usepackage{graphicx}
\usepackage{amsmath}
\usepackage{amssymb}

\usepackage{multirow}
\usepackage{subfigure}
\usepackage{bm}
\usepackage{color}
\usepackage[breaklinks=true,bookmarks=false]{hyperref}

\cvprfinalcopy % *** Uncomment this line for the final submission

\def\cvprPaperID{9265} % *** Enter the CVPR Paper ID here

\ifcvprfinal\fi
\begin{document}

%%%%%%%%% TITLE
\title{Transferring and Regularizing Prediction for Semantic Segmentation\thanks{{\small This work was performed at JD AI Research.}}}

\author{Yiheng Zhang$^{\dag}$, Zhaofan Qiu$^{\dag}$, Ting Yao$^{\ddag}$, Chong-Wah Ngo$^{\S}$, Dong Liu$^{\dag}$, and Tao Mei$^{\ddag}$ \\
{\normalsize $^{\dag}$ University of Science and Technology of China, Hefei, China}\\
{\normalsize $^{\ddag}$ JD AI Research, Beijing, China~~~~~~~~$^{\S}$ City University of Hong Kong, Kowloon, Hong Kong}\\
{\tt\small \{yihengzhang.chn, zhaofanqiu, tingyao.ustc\}@gmail.com}\\
{\tt\small cscwngo@cityu.edu.hk, dongeliu@ustc.edu.cn, tmei@jd.com}
}

\maketitle
\thispagestyle{empty}

%%%%%%%%% ABSTRACT
\begin{abstract}
   \vspace{-0.1in}
   Semantic segmentation often requires a large set of images with pixel-level annotations. In the view of extremely expensive expert labeling, recent research has shown that the models trained on photo-realistic synthetic data (e.g., computer games) with computer-generated annotations can be adapted to real images. Despite this progress, without constraining the prediction on real images, the models will easily overfit on synthetic data due to severe domain mismatch.
   In this paper, we novelly exploit the intrinsic properties of semantic segmentation to alleviate such problem for model transfer.
   Specifically, we present a Regularizer of Prediction Transfer (RPT) that imposes the intrinsic properties as constraints to regularize model transfer in an unsupervised fashion. These constraints include patch-level, cluster-level and context-level semantic prediction consistencies at different levels of image formation. As the transfer is label-free and data-driven, the robustness of prediction is addressed by selectively involving a subset of image regions for model regularization. Extensive experiments are conducted to verify the proposal of RPT on the transfer of models trained on GTA5 and SYNTHIA (synthetic data) to Cityscapes dataset (urban street scenes). RPT shows consistent improvements when injecting the constraints on several neural networks for semantic segmentation. More remarkably, when integrating RPT into the adversarial-based segmentation framework, we report to-date the best results: mIoU of 53.2\%/51.7\% when transferring from GTA5/SYNTHIA to Cityscapes, respectively.
\end{abstract}
%%%%%%%%% BODY TEXT
\vspace{-0.15in}
\section{Introduction}
\vspace{-0.05in}
Semantic segmentation aims at assigning semantic labels to every pixel of an image. Leveraging on CNNs~\cite{he2016deep,Hu_2018_CVPR,ILSVRC15,simonyan2014very,szegedy2015going}, significant progress has been reported for this fundamental task~\cite{chen2016deeplab,Chen_2018_ECCV,long2015fully,Peng_2017_CVPR}. One drawback of the existing approaches, nevertheless, is the requirement of large quantities of pixel-level annotations, such as in VOC \cite{everingham2010pascal}, COCO \cite{lin2014microsoft} and Cityscapes \cite{Cordts2016Cityscapes} datasets, for model training. Labeling of semantics at pixel-level is cost expensive and time consuming.
For example, the Cityscapes dataset is composed of 5,000 high-quality pixel-wise annotated images, and the annotation on a single image is reported to take more than 1.5 hours.

An alternative is by utilizing synthetic data, which is largely available in 3D engines (e.g., SYNTHIA \cite{ros2016synthia}) and 3D computer games (e.g., GTA5 \cite{GTA5_richter2016playing}). The ground-truth semantics of these data can be automatically generated without manual labeling. Nevertheless, in the case where the synthetic data is different from the real images, the domain gap might be difficult to bridge. Unsupervised domain adaptation is generally regarded as an appealing way to address the problem of domain gap. The existing approaches include narrowing the gap by transferring images across domains \cite{dundar2018domain,murez2018image,wu2018dcan} and learning domain-invariant representation via adversarial mechanism \cite{Du_2019_ICCV,luo2019taking,Vu_2019_CVPR}.

\begin{figure*}[!tb]
\vspace{-0.05in}
   \centering {\includegraphics[width=0.98\textwidth]{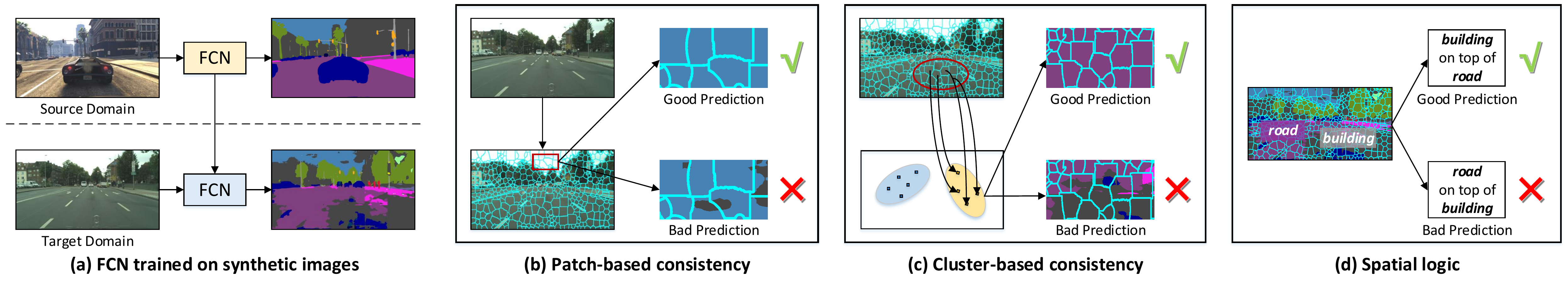}}
   \vspace{-0.1in}
   \caption{\small The examples of (a) predictions on two domains by fully convolutional networks trained on synthetic data; (b)$\sim$(d) the three evaluation criteria we studied, i.e., patch-based consistency, cluster-based consistency and spatial logic.}
   \label{fig:intro}
   \vspace{-0.25in}
\end{figure*}

In this paper, we consider model overfitting in source domain as the major cause of domain mismatch. As shown in Figure \ref{fig:intro}(a), although Fully Convolutional Networks (FCN) perfectly segment the synthetic image by correct labeling of pixels, directly deploying this model for real image yields poor results. Instead of leveraging training samples in the target domain for model fine-tuning, this paper explores label-free constraints to alleviate the problem of model overfitting. These constraints are intrinsic and generic in the context of semantic segmentation. Figure \ref{fig:intro}(b)$\sim$(d) illustrate three label-free constraints being investigated. The first two constraints, namely patch-based and cluster-based consistencies guide the segmentation based on the prediction consistency among the pixels in an image patch and among the clusters of patches sharing similar visual properties, respectively. The last criterion, namely spatial logic, contextualizes the prediction of labels based on spatial relation between image patches. Based on these criteria, we propose a novel Regularizer of Prediction Transfer (RPT) for transferring the model trained on synthetic data for semantic segmentation of real images.

The main contribution of this paper is on the exploration of label-free data-driven constraints for transferring of model to bridge domain gap. These constraints are imposed as regularizers during training to transfer an overfitted source model for proper labeling of pixels in the target domain. Specifically, at the lowest level of regularization, majority voting is performed to derive a dominative category for each image patch. The dominative category serves as a local cue for pixels with low prediction confidence to adjust their label prediction during training. The patch-level regularization is then extended to a higher level of regularization to explore cluster-level and context-level prediction consistency.
Despite its simplicity, the three regularizers, when jointed optimized in a fully convolutional network with adversarial learning, show impressive performances by outperforming several state-of-the-art methods, when transferring the models trained on GTA5 and SYNTHIA for semantic segmentation on the Cityscapes dataset.

\vspace{-0.05in}
\section{Related Work}
\vspace{-0.05in}
\textbf{CNN Based Semantic Segmentation.} As one of the most challenging computer vision task, semantic segmentation has received intensive research attention. With the surge of deep learning and convolutional neural networks (CNNs), Fully Convolutional Network (FCN)~\cite{long2015fully} successfully serves as an effective approach that employs CNNs to perform dense semantic prediction. Following FCN, various schemes, ranging from multi-path feature aggregation and refinement~\cite{ghiasi2016laplacian,Lin:2017:RefineNet,Peng_2017_CVPR,Pohlen_2017_CVPR,zhang2019customizable,Zhao_2018_ECCV} to multi-scale context extraction and integration~\cite{chen2018searching,chen2016deeplab,He_2019_ICCV,qiu2017learning,yang2018denseaspp,Zhang_2018_CVPR_context,zhao2017pspnet}, have been developed and achieved great success in leveraging contextual information for semantic segmentation. Post-processing techniques, such as CRF~\cite{chen2016deeplab} and MRF~\cite{liu2018deep}, could further be applied to take the spatial consistency of labels into account and improve the predictions from FCNs. Considering that such methods typically rely on the datasets with pixel-level annotations which are extremely expensive and laborious to collect, researchers have also strived to utilize a weaker form of annotation, such as image-level tags \cite{papandreou2015weakly,pinheiro2015image}, bounding boxes~\cite{dai2015boxsup}, scribbles~\cite{bearman2016s} and statistics \cite{pathak2015constrained}, for semantic segmentation. The development of computer graphics techniques provides an alternative approach that exploits synthetic data with free annotations. This work aims to study the methods of applying the semantic segmentation model learnt on the computer-generated synthetic data to unlabeled real data.

\textbf{Domain Adaptation of Semantic Segmentation.}
To alleviate the issues of expensive labeling efforts in collecting pixel-level annotations, domain adaptation is studied for semantic segmentation. FCNWild~\cite{BDDS_hoffman2016fcns}, which is one of the early works, attempts to align the features in different domains from both global and local aspects by adversarial training. Curriculum~\cite{zhang2017curriculum} proposes a curriculum-style learning approach to bridge the domain gap between synthetic and real data. Later on, similar to domain adaptation in image recognition and object detection \cite{cai2019exploring,pan2019transferrable,yao2015semi}, visual appearance-level and/or representation-level adaptation are exploited in~\cite{dundar2018domain,murez2018image,Tsai_2018_CVPR,Zhang_2018_CVPR} for this task. \cite{dundar2018domain,murez2018image} perform an image-to-image translation that transfers the synthetic images to the real domain in the appearance-level. From the perspective of the representation-level adaptation, AdaSegNet~\cite{Tsai_2018_CVPR} proposes to apply adversarial learning on segmentation maps for adapting structured output space.
FCAN~\cite{Zhang_2018_CVPR} employs the two levels of adaptation simultaneously, in which the appearance gap between synthetic and real images is minimized and the network is encouraged to learn domain-invariant representations.
There have been several other strategies~\cite{chang2019all,chen2019learning,chen2018road,pmlr-v80-hoffman18a,iqbal2019mlsl,li2019bidirectional,zou2018unsupervised}, being performed for cross-domain semantic segmentation.
For example, ROAD~\cite{chen2018road} devises a target guided distillation module and a spatial-aware adaptation module for real style and distribution orientation. Labels from the source domain are transferred to the target domain as the additional supervision in CyCADA~\cite{pmlr-v80-hoffman18a}. Depth maps which are available in virtual 3D environments are utilized as geometric information to reduce domain shift in ~\cite{chen2019learning}. \cite{iqbal2019mlsl,li2019bidirectional,zou2018unsupervised} treat target predictions as the guide for learning a model applicable to the images in target domain by self-supervised learning. \cite{chang2019all} proposes a domain invariant structure extraction framework that decouples the structure and texture representations of images and improves the performance of segmentation.

\textbf{Summary.} Most of the aforementioned approaches mainly investigate the problem of domain adaptation for semantic segmentation through bridging the domain gap during training. Our work is different in the way that we seek the additional regularization for the prediction in target domain based on the intrinsic and generic properties of semantic segmentation task. Such solution formulates an innovative and promising research direction for this task.

\begin{figure}[!tb]
   \centering {\includegraphics[width=0.478\textwidth]{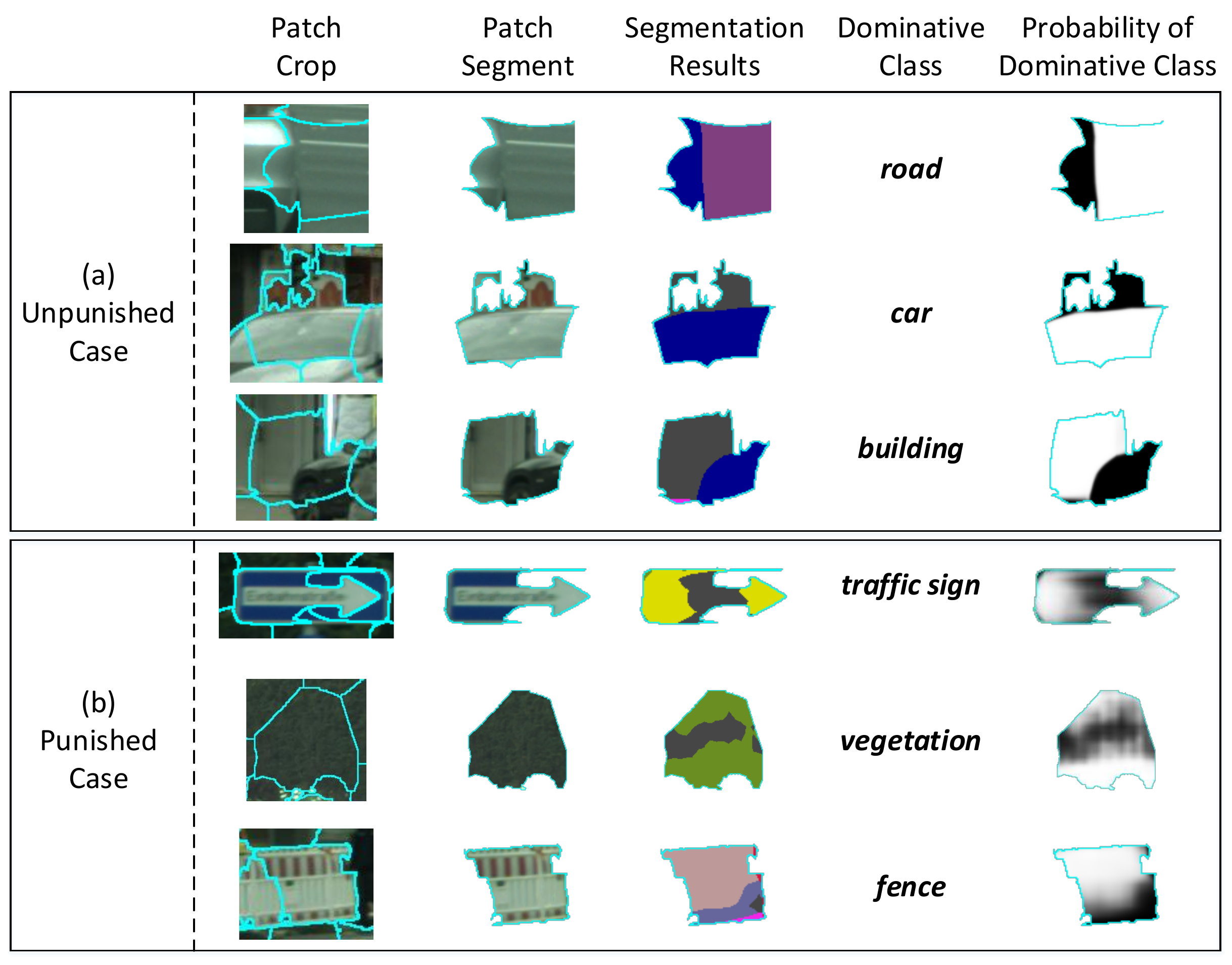}}
   \caption{\small Example of pixels to be unpunished (a) or punished (b) in optimization. (a) For the unpunished cases, some pixels are very confident in the class differed from the dominative category. (b) For the punished cases, most pixels inside the region predict relatively high probabilities for the dominative category.}
   \label{fig:patch}
   \vspace{-0.15in}
\end{figure}
\section{Regularizer of Prediction Transfer}
We start by introducing the Regularizer of Prediction Transfer (RPT) for semantic segmentation.
Three criteria are defined to assess the quality of segmentation. The result of assessment is leveraged to guide the transfer of a learnt model in the source domain for semantic segmentation in the target domain.

\subsection{Patch-based Consistency}
The idea is to enforce all pixels in a patch to be consistent in the prediction of semantic labels. Here, a patch is defined as a superpixel that groups neighboring pixels with similar visual appearance. We employ Simple Linear Iterative Clustering (SLIC)~\cite{achanta2012slic}, which is both speed and memory efficient in the generation of superpixels by adopting k-means algorithm.
Given one image from target domain $x_t$, SLIC splits the image into $N$ superpixels $\{S_i|i=1,...,N\}$. Each superpixel $S_i=\{p^j_i|j=1,...,M_i\}$ is composed of $M_i$ adjacent pixels with similar appearance.
We assume that all or the majority of pixels will be annotated with the same semantic labels. Here, the dominative category $\hat{y}_i$ of a superpixel is defined as the most number of predicted labels among all the pixels in this superpixel.

As SLIC considers only visual cue, a superpixel usually contains multiple regions of different semantic labels. Simply involving all pixels in network optimization can run into the risk of skew optimization. To address this problem, a subset of pixels is masked out from patch-based regularization.
Specifically, in superpixel $S_i$, pixels $p_i^j\in S_i$ are clustered into two groups depending on the predicted probability of the dominative category $\hat{y}_i$: (a) $P_{seg}(\hat{y}_i| p^j_i)<=\lambda_{pc}$ means that the probability is less than or equal to a pre-defined threshold $\lambda_{pc}$. In other words, the pixel $p_i^j$ is predicted with labels different from the dominative category with relatively high probability. This group of pixels should be exempted from regularization. (b) $P_{seg}(\hat{y}_i| p^j_i)>\lambda_{pc}$ represents that $p_i^j$ has relatively higher confidence to be predicted as the dominative category. In this case, the dominative $\hat{y}_i$ is leveraged as a cue to guide the prediction of these pixels. To the end, the loss item for patch-based consistency regularization of a target image $x_t$ is formulated as:
\begin{equation}\label{eq:pc}
\begin{aligned}
\mathcal{L}_{pc}(x_t)=- \sum_{i, j} I_{(P_{seg}(\hat{y}_i| p^j_i)>\lambda_{pc})} log P_{seg}(\hat{y}_i| p^j_i)
\end{aligned}~~,
\end{equation}
where $I_{(\cdot)}$ is an indicator function to selectively mask out pixels from optimization by thresholding. Figure \ref{fig:patch} shows examples of superpixels that are masked out (i.e., unpunished) and involved (i.e., punished) for optimization.

\subsection{Cluster-based Consistency}

\begin{figure}[!tb]
   \centering {\includegraphics[width=0.40\textwidth]{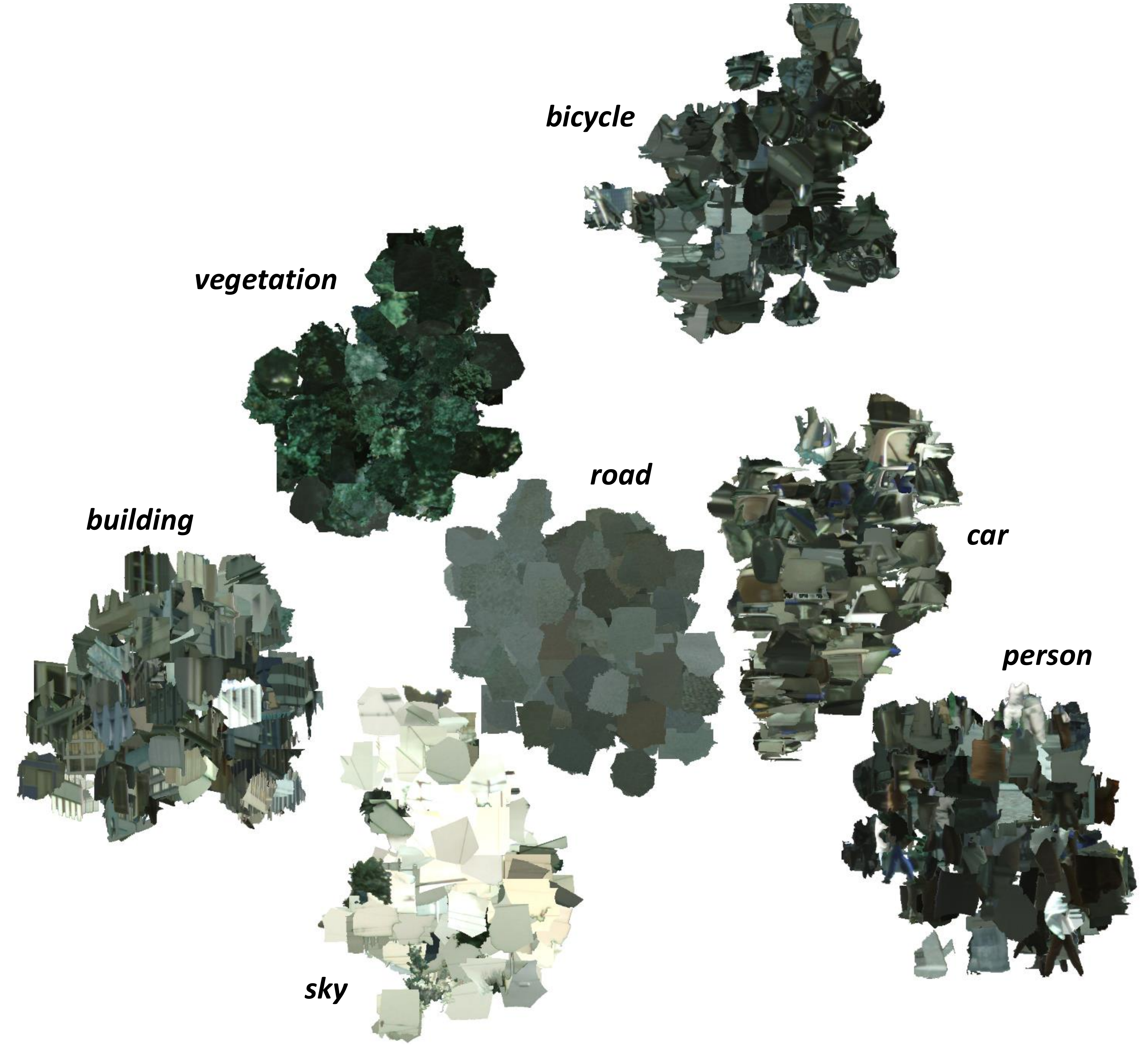}}
   \caption{\small Feature space visualization of seven superpixel clusters using t-SNE. The dominative category is given for each cluster.}
   \label{fig:cluster}
   \vspace{-0.15in}
\end{figure}
In addition to patch, we also enforce the consistency of label prediction among the clusters of patches that are visually similar. Specifically, cluster-level regularization imposes a constraint that the superpixels with similar visual properties should predict the cluster dominative category as their label. To this end, superpixels are further grouped into clusters. The feature representation of a superpixel is extracted through ResNet-101~\cite{he2016deep}, which is pre-trained on ImageNet dataset~\cite{ILSVRC15}. The feature vector utilized for clustering is generated by averagely pooling the feature maps of the superpixel region from $res5c$ layer.
All the superpixels from target domain images are grouped into $K=2048$ clusters by k-means algorithm. The cluster-level dominative category $\tilde{y}_k$ is determined by majority voting among the superpixels within a cluster. Figure \ref{fig:cluster} visualizes seven examples of clusters and the corresponding dominative categories by t-SNE \cite{maaten:JMLR08}. As clustering is imperfect, it is expected that some superpixels will be incorrectly grouped. Denote $P_{seg}(\tilde{y}_k| p^j_i)$, where $p^j_i \in S_i \in C_k$, as the probability of predicting cluster-level dominative category as label for pixel $p^j_i$. Similar to patch-based consistency regularization, pixels with low confidence on the cluster-level category will not be punished during network optimization. Thus, the loss item of cluster-based consistency regularization for a target image $x_t$ is defined as:
\begin{equation}\label{eq:pc}
\begin{aligned}
\mathcal{L}_{cc}(x_t)=- \sum_{i, j, S_i \in C_k} I_{(P_{seg}(\tilde{y}_k| p^j_i)>\lambda_{cc})} log P_{seg}(\tilde{y}_k| p^j_i)
\end{aligned}~~,
\end{equation}
where $\lambda_{cc}$ is a pre-defined threshold to gate whether a pixel should be masked out from regularization.

\subsection{Spatial Logic}
A useful cue to leverage for target-domain segmentation is the spatial relation between semantic labels. For instance, a superpixel of category \emph{sky} is likely on the top of another superpixel labeled with \emph{building} or \emph{road}, and not vice versa. These relations are expected to be invariant across the source and target domains. The supportive hypothesis behind is introduced in \cite{chang2019all} that the high-level structure information of an image is informative for semantic segmentation and can be readily shared across domains. As such, the motivation of spatial logic is to preserve the spatial relations learnt in source domain to target domain.

\begin{figure}[!tb]
   \centering {\includegraphics[width=0.45\textwidth]{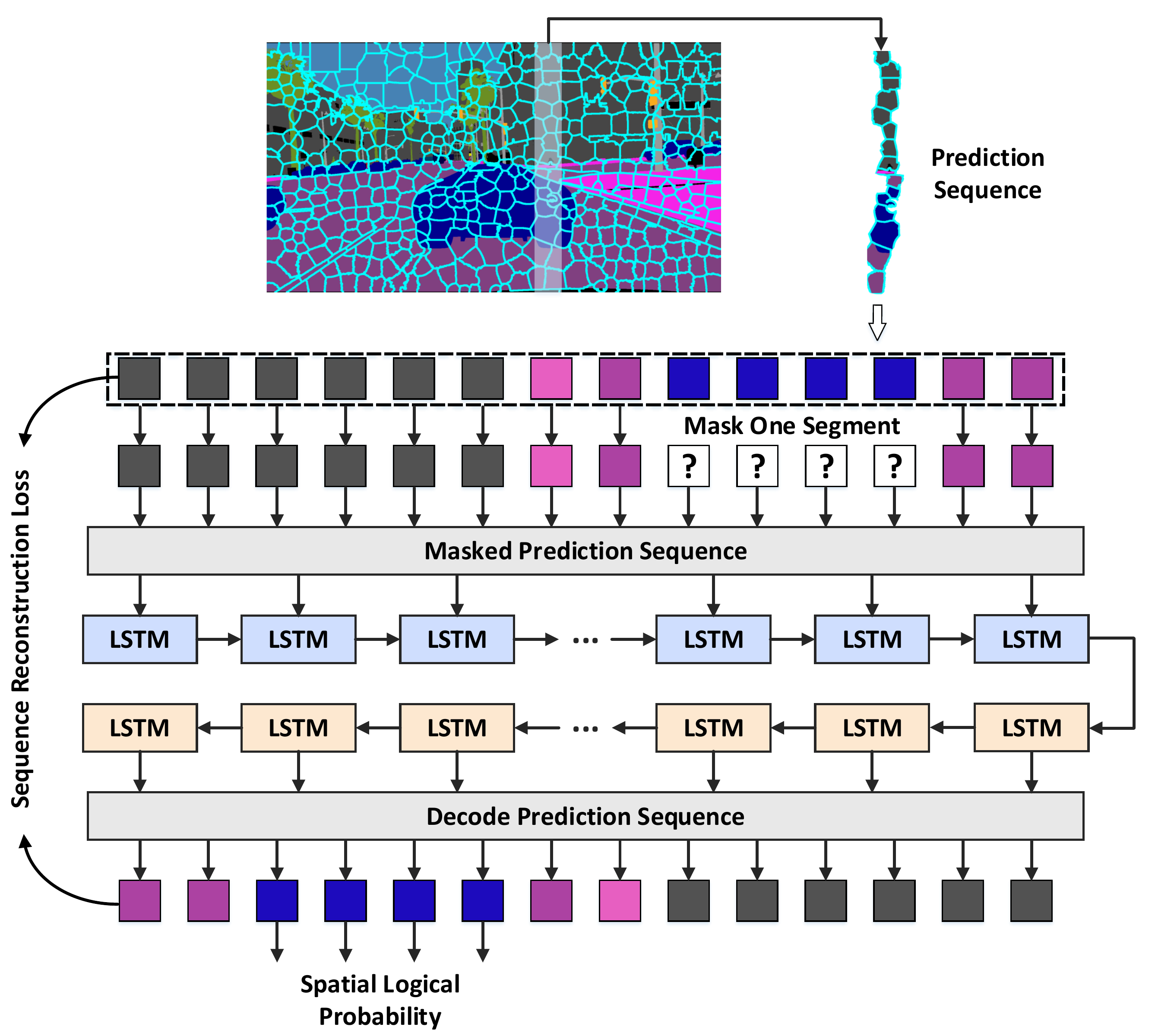}}
   \caption{\small The LSTM encoder-decoder architecture to learn the spatial logic in the prediction map.}
   \label{fig:spatial}
   \vspace{-0.15in}
\end{figure}

Formally, we exploit the LSTM encoder-decoder architecture to learn the vertical relation between superpixels, as shown in Figure \ref{fig:spatial}. The main goal of this architecture is to speculate the category of the masked segment in the sequence according to context information. Then, the produced probability can be used to evaluate the logical validity of the predicted category in the masked segment. Suppose we have a prediction sequence $\mathcal{Y}$, where $\mathcal{Y}=\{\mathbf{y}_1,\mathbf{y}_2,...,\mathbf{y}_{T-1},\mathbf{y}_{T}\}$ including $T$ superpixel predictions sliced from one column of prediction map. Let $\mathbf{y}_{t} \in \mathbb{R}^{C+1}$ denote the one-hot vector of the $t$-th prediction in the sequence, and the dimension of $\mathbf{y}_{t}$, i.e., $C+1$, is the number of semantic categories plus one symbol as an identification of masked prediction. The masked prediction sequence $\hat{\mathcal{Y}}$, which is fed into the LSTM encoder, is generated by masking a segment of consecutive predictions with the identical semantic category in the original sequence $\mathcal{Y}$. The LSTM encoder embeds the masked prediction sequence $\hat{\mathcal{Y}}$ into a sequence representation. The LSTM decoder, which is attached on the top of the encoder, then speculates the categories of the masked segment and reconstructs the original sequence $\mathcal{Y}$. To learn the aforementioned spatial logic, the encoder-decoder architecture is optimized with the cross-entropy loss supervised by the label from source domain.

Next, the optimized model can be utilized to estimate the validity of each prediction from the view of spatial logic. For the target image $x_t$, we first slice the prediction map to several columns consisting of vertically neighbored superpixels. The patch-level dominative categories of the superpixels in the column are organized into a prediction sequence. For the superpixel $S_i$ in the column, the spatial logical probability $P_{logic}(\hat{y}_i| S_i)$ is measured by the LSTM encoder-decoder only when the prediction of this superpixel is masked in the input sequence. Once this probability is lower than the threshold $\lambda_{sl}$, we consider this prediction to be illogical and punish the prediction of $\hat{y}_i$ by the segmentation network. The loss of spatial logic regularization is computed as:
\begin{equation}\label{eq:pc}
\begin{aligned}
\mathcal{L}_{sl}(x_t)=\sum_{i, j} I_{(P_{logic}(\hat{y}_i| S_i)<\lambda_{sl})} log P_{seg}(\hat{y}_i| p^j_i)
\end{aligned}~~,
\end{equation}
where $P_{logic}(\cdot)$ denotes the prediction from LSTM encoder-decoder architecture.

\section{Semantic Segmentation with RPT}
\begin{figure}[!tb]
   \centering {\includegraphics[width=0.45\textwidth]{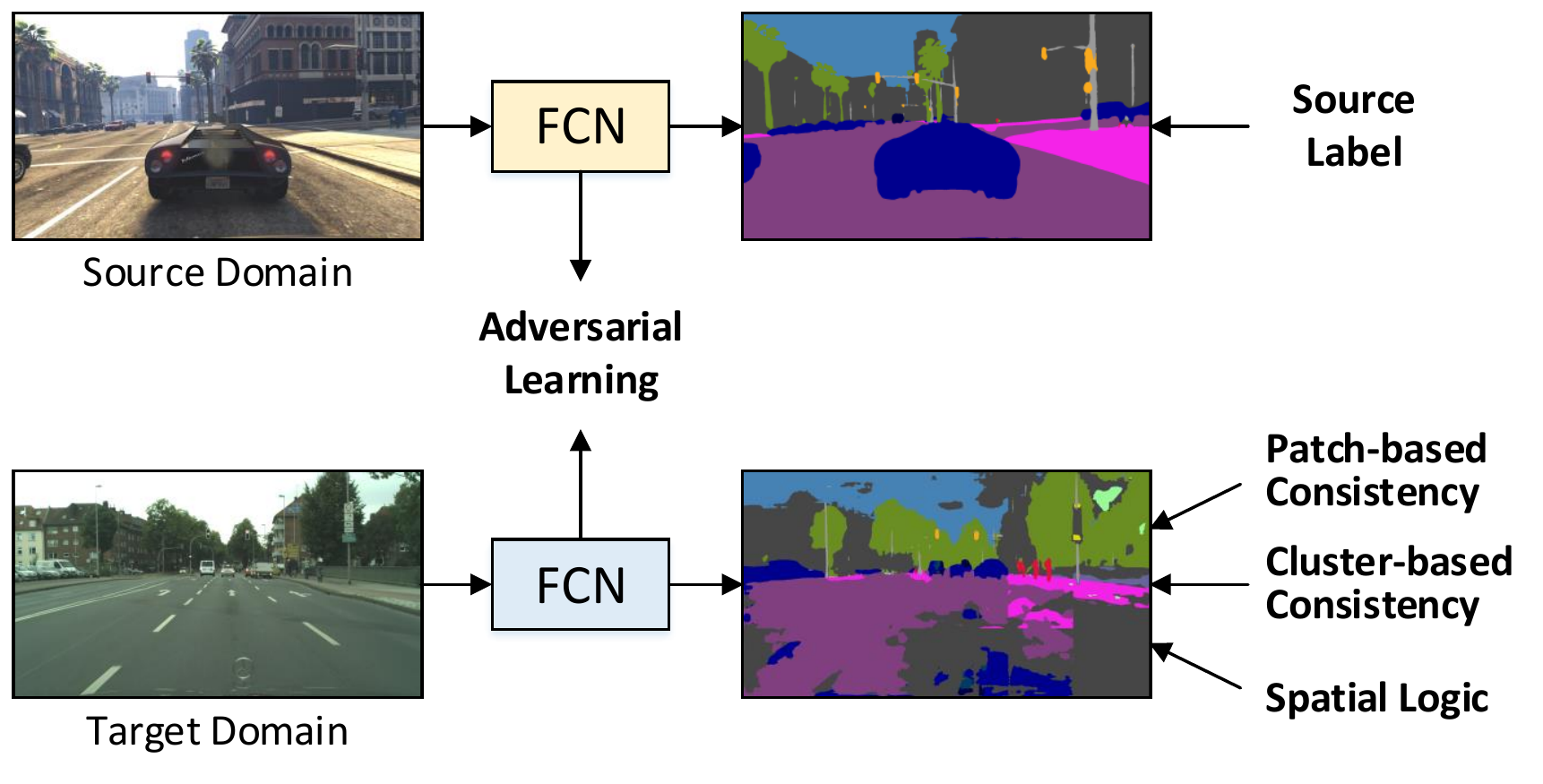}}
   \vspace{-0.05in}
   \caption{\small The adversarial-based semantic segmentation adaptation framework with RPT. The shared FCN is learnt with adversarial loss for domain-invariant representations across two domains. The predictions on source domain are optimized by supervised label, while the target domain predictions are regularized by RPT loss.}
   \label{fig:framework}
   \vspace{-0.15in}
\end{figure}

The proposed Regularizer of Prediction Transfer (RPT) can be easily integrated into most of the existing frameworks for domain adaptation of semantic segmentation. Here, we choose the widely adopted framework based on adversarial learning as shown in Figure \ref{fig:framework}. The principle in this framework is equivalent to guiding the semantic segmentation in both domains by fooling a domain discriminator $D$ with the learnt source and target representations. Formally, given the training set $\mathcal{X}_{s}=\{x_s^{i}|i=1,\dots,N_s\}$ in source domain and $\mathcal{X}_{t}=\{x_t^{i}|i=1,\dots,N_t\}$ in target domain, the adversarial loss $\mathcal{L}_{adv}$ is the average classification loss, which is formulated as:
\begin{equation}
   \label{eq:adv}
   \begin{aligned}
   \mathcal{L}_{adv}(\mathcal{X}_{s},\mathcal{X}_{t})= \mathop{-E}\limits_{x_t \sim \mathcal{X}_t}[log(D(x_t))]\mathop{-E}\limits_{x_s \sim \mathcal{X}_s}[log(1 - D(x_s)]
   \end{aligned}~~.
\end{equation}
where $\mathop{E}$ denotes the expectation over the image set. The discriminator $D$ will attempt to minimize this loss by differentiating between source and target representations, and the shared Fully Convolutional Network (FCN) is learnt to fool the domain discriminator.
Considering that the image region corresponding to the receptive field of each spatial unit in the final feature map is treated as an individual instance during semantic segmentation, the representations of such instances are expected to be invariant across domains.
Thus we employ a fully convolutional domain discriminator whose outputs are the domain prediction of each image region corresponding to the spatial unit in the feature map.

Since training labels are available in the source domain, the loss function is based on the pixel-level classification loss $\mathcal{L}_{seg}$. In contrast, due to the absence of training labels, the loss function in the target domain is defined based upon the following three regularizers:
\begin{equation}
   \label{eq:rpt}
   \small
   \begin{aligned}
   \mathcal{L}_{rpt}(\mathcal{X}_{t})= \mathop{E}\limits_{x_t \sim \mathcal{X}_t}[\mathcal{L}_{cc}(x_t)+\mathcal{L}_{pc}(x_t)+\mathcal{L}_{sl}(x_t)]
   \end{aligned}~~.
\end{equation}
Here, we empirically treat each loss in RPT equally. Thus, the overall objective of the segmentation framework integrates $\mathcal{L}_{adv}$, $\mathcal{L}_{seg}$ and $\mathcal{L}_{rpt}$ as:
\begin{equation}
   \label{eq:all}
   \small
   \begin{aligned}
   \mathop{\min}_{FCN}\{-\varepsilon \mathop{\min}_{D}\mathcal{L}_{adv}(\mathcal{X}_s, \mathcal{X}_t) + \mathcal{L}_{seg}(\mathcal{X}_s) + \mathcal{L}_{rpt}(\mathcal{X}_t)\}
   \end{aligned}~~,
\end{equation}
where $\varepsilon=0.1$ is the trade-off parameter to align the scale of different losses.

\section{Implementation} \label{sec:imp}
\textbf{Training strategy.} Our proposed network is implemented in Caffe~\cite{jia2014caffe} framework and the weights are trained by SGD optimizer. We employ dilated FCN~\cite{chen2016deeplab} originated from the ImageNet pre-trained ResNet-101 as our backbone followed by a PSP module~\cite{zhao2017pspnet}, unless otherwise stated. The domain discriminator for adversarial learning is borrowed from FCAN~\cite{Zhang_2018_CVPR}. During the training stage, images are randomly cropped to $713\times713$ due to the limitation of GPU memory. Both random horizontal flipping and image resizing are utilized for data augmentation. To make the training process stable, we pre-train the FCN on data from the source domain with annotations. At the stage of pre-training, the ``poly'' policy whose power is fixed to 0.9 is adopted with the initial learning rate 0.001. Momentum and weight decay are 0.9 and 0.0005 respectively. Each mini-batch has 8 samples and maximum training iterations is set as 30K. With the source domain pre-trained weights, we perform the domain adaptation by finetuning the whole adaptation framework which is equipped with our proposed RPT. The initial learning rate is 0.0001 and the total training iteration is 10K. Other training hyper-parameters remain unchanged.
Following \cite{lian2019constructing}, we randomly selected 500 images from the official training set of Cityscapes as a general validation set. The hyper-parameters ($\lambda_{pc}=\lambda_{cc}=\lambda_{sl}=0.25$, $\varepsilon=0.1$) are all determined on this set.

\textbf{Complexity of superpixel.}
RPT highly relies on the quality of superpixel extraction. For robustness, superpixels with complex content ideally should be excluded from model training. The term ``complex'' refers to the distribution of semantic labels in a superpixel. In our case, we measure complexity based on the proportion of pixels being predicted with the dominative category over the number of pixels in a superpixel. A larger value implies consistency in prediction and hence safer to involve the corresponding superpixel in regularizations. Empirically, RPT only regularizes the top-50\% of superpixels. The empirical choice will be further validated in the next section.

\textbf{State update of RPT.}
During network optimization, the segmentation prediction $P_{seg}$, superpixel dominative category $\hat{y}_i$ and cluster dominative category $\tilde{y}_k$ change gradually. Iteratively updating these ``states'' is computationally expensive because reassigning the categories to superpixel and cluster (e.g., $\hat{y}_i$ and $\tilde{y}_k$) requires the semantic predictions collected from the whole training set of the target domain. Considering these predictions only change slightly during training, we first calculate these states before the optimization (without regularization) and fix these states at the beginning of iterations. Then, we will update the predictions or states for $N_{su}$ times evenly during training.

\vspace{-0.1in}
\section{Experiments}
\vspace{-0.05in}
\subsection{Datasets}
\vspace{-0.05in}
The experiments are conducted on GTA5~\cite{GTA5_richter2016playing}, SYNTHIA~\cite{ros2016synthia} and Cityscapes~\cite{Cordts2016Cityscapes} datasets. The proposed RPT is trained on GTA5 and SYNTHIA (source domain) and Cityscapes (target domain). GTA5 is composed of 24,966 synthetic images of size $1914 \times 1052$. These images are generated by Grand Theft Auto V (GTA5), a modern computer game, to render city scenes. The pixels of these images are annotated with 19 classes that are compatible with the labels in Cityscapes. Similarly, SYNTHIA consists of synthetic images of urban scenes with resolutions of $1280 \times 760$. Following~\cite{chang2019all,chen2019learning,hong2018conditional,li2019bidirectional,Tsai_2018_CVPR}, we use the subset, SYNTHIA-RAND-CITYSCAPES, which has 9,400 images being annotated with labels consistent with Cityscapes for experiments. Cityscapes is composed of 5,000 images of resolution $2048 \times 1024$. These images are split into three subsets of sizes 2,975, 500 and 1,525 for training, validation and testing, respectively. The pixels of these images are annotated with 19 classes. In the experiments, the training subset is treated as the target-domain training data, where the pixel-level annotation is assumed unknown to RPT. On the other hand, the target-domain testing data is from validation subset. The same setting is also exploited in~\cite{chang2019all,li2019bidirectional,Tsai_2018_CVPR}.

To this end, the performance of RPT is assessed by treating GTA5 as source domain and Cityscapes as target domain (i.e., GTA5~$\to$~Cityscapes), and similarly, SYNTHIA~$\to$~Cityscapes. The metrics are per class Intersection over Union (IoU) and mean IoU over all the classes.

\begin{table}
   \centering
   \small
   \caption{\small RPT performances in terms of mean IoU for domain adaptation of semantic segmentation on GTA5~$\to$~Cityscapes.}
   \begin{tabular}{l|c@{~~}c@{~~}c|c@{~~}c@{~~}c} \hline
      \multirow{2}{*}{\textbf{Method}}  & \multicolumn{3}{c|}{\textbf{ResNet-50}}& \multicolumn{3}{c}{\textbf{ResNet-101}}\\
      & FCN & +ABN & +ADV & FCN & +ABN & +ADV \\ \hline
      baseline & 30.1 & 35.7 & 45.7 & 32.3 & 39.1 & 47.2 \\ \hline
      \textbf{RPT$^{1}$} & 33.0 & 39.3 & 48.7 & 36.1 & 42.9 & 50.4 \\
      \textbf{RPT$^{2}$} &  33.4 & 39.9 & 50.0 & 37.9 & 44.2 & 51.7\\
      \textbf{RPT$^{3}$} & 33.5 & 40.0 & 50.0 & 39.1 & 44.6 & 52.6\\ \hline
   \end{tabular}
   \label{tab:effectiveness}
   \vspace{-0.15in}
\end{table}

\begin{figure}[!tb]
   \centering
   \subfigure[State updating]{
      \label{fig:curve:a}
      \includegraphics[width=0.22\textwidth]{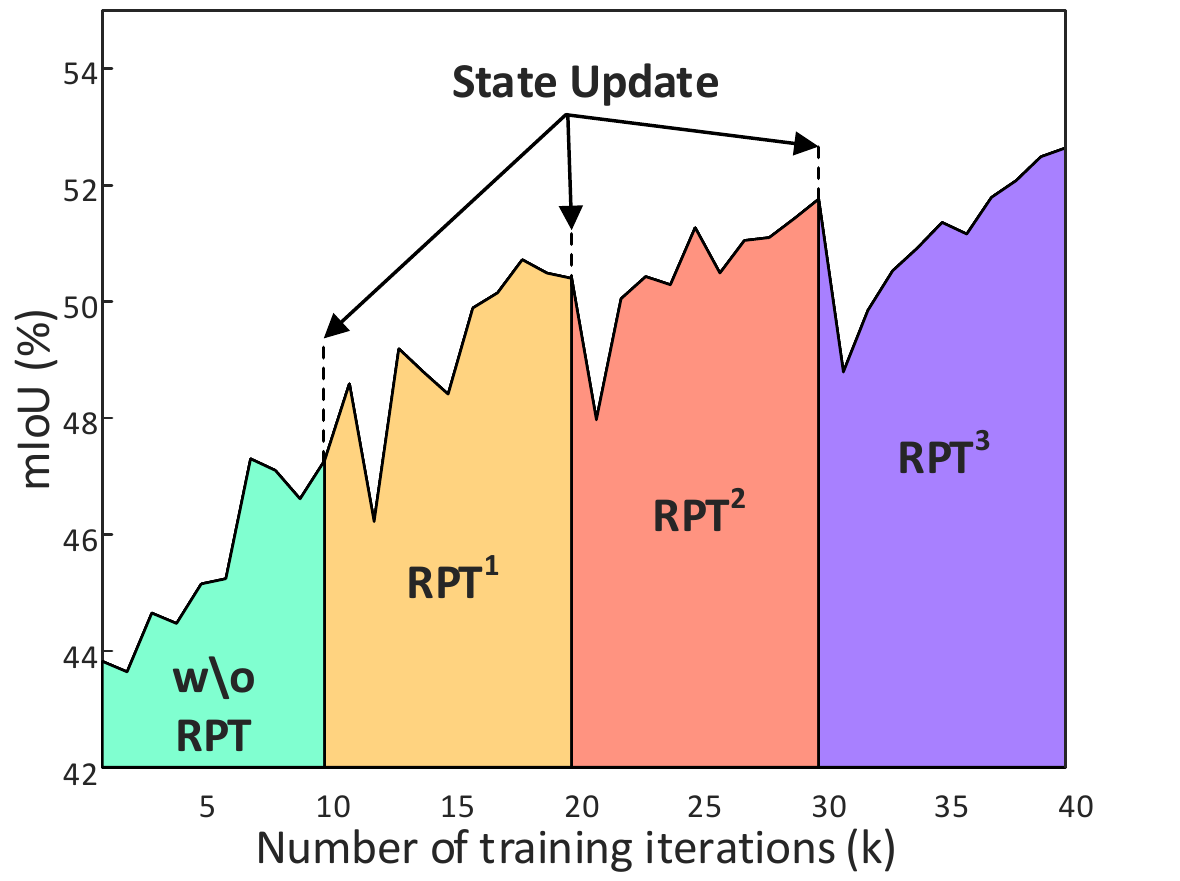}}
   \subfigure[Filtering complex superpixels]{
      \label{fig:curve:b}
      \includegraphics[width=0.23\textwidth]{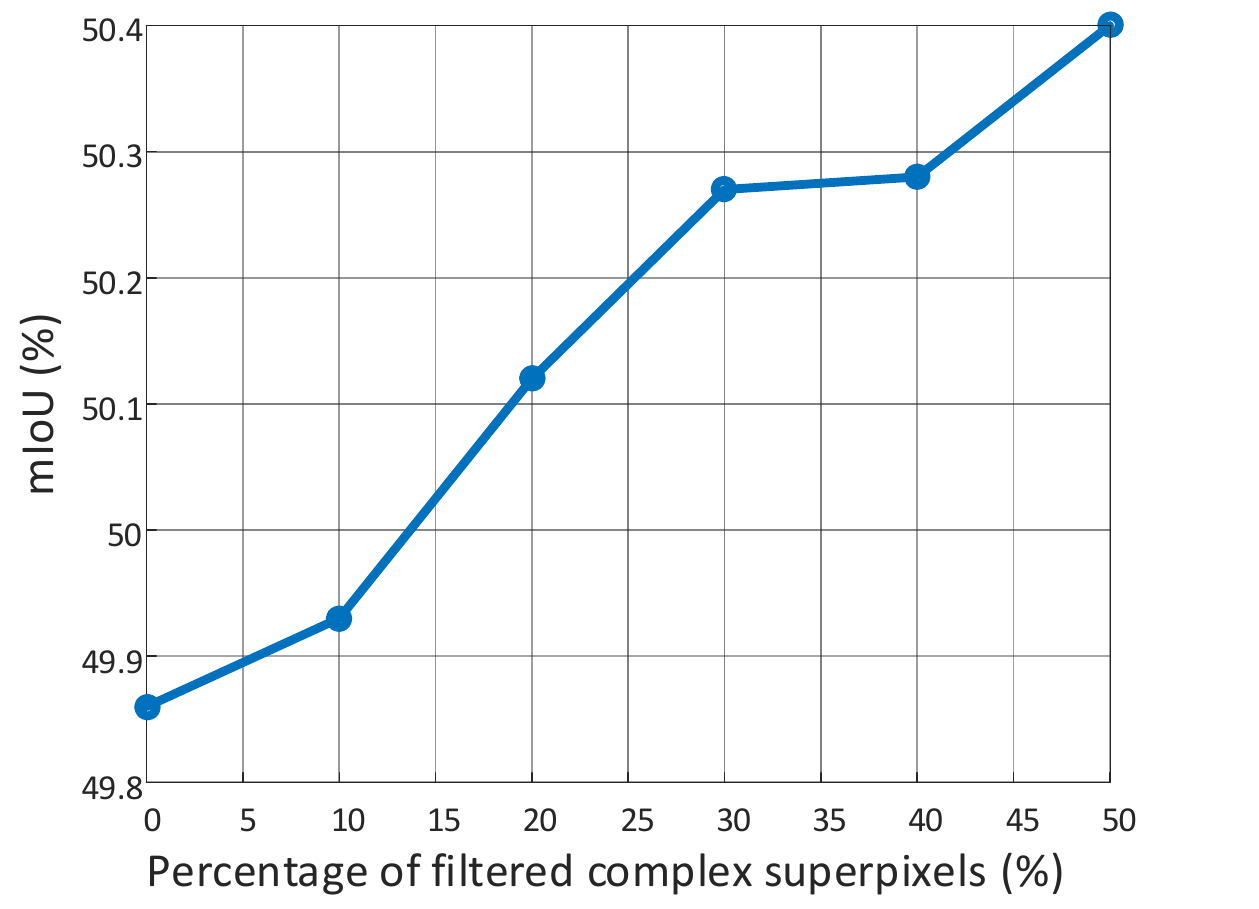}}
   \caption{\small Two analysis experiments of (a) the effectiveness of state updating during training of RPT$^{3}$; (b) the percentage of filtered complex superpixels of RPT$^{1}$.}
   \label{fig:curve}
   \vspace{-0.15in}
\end{figure}
\subsection{Evaluation of RPT}
RPT is experimented on top of six different network architectures derived from \textbf{FCN} which leverages on either \textbf{ResNet-50} or \textbf{ResNet-101} as the backbone network. Specially, we adopt Adaptive Batch Normalization (\textbf{ABN}) to replace the mean and variance of BN in the original version of FCN, resulting in a variant of network named FCN+ABN. Note that the BN layer is first learnt in source domain and then replaced by ABN when being applied to the target domain. In addition, leveraging on the adversarial training (\textbf{ADV}), another variant, FCN+ABN+ADV, is trained to learn domain-invariant representations.

We first verify the impact of $N_{su}$, the number of state updating, in RPT. Table~\ref{tab:effectiveness} summarizes the impact on six variants of network for domain adaptation on GTA5~$\to$~Cityscapes.
All the networks are pre-trained on ImageNet dataset and then injected with RPT.
The superscript, RPT$^{n}$, refers to the number of times for state updating (see Table~\ref{tab:effectiveness} for exact number).
The baselines are obtained by performing domain adaptation of semantic segmentation on the use of the corresponding network architectures, but without RPT.
Overall, RPT improves the baseline without regularization. The improvement is consistently observed across the variants of networks, and proportional to the number of state updating at the expense of computation cost. RPT$^{3}$ achieves the best performance (mIoU = 52.6\%) and with 5.4\% improvement over the baseline of the same network (FCN+ABN+ADV). Figure \ref{fig:curve:a} shows the performance changes in terms of mIoU during training over different times of state updating. The training starts with model learning in source domain. State updating, such as the assignment of dominative categories at superpixel and cluster levels, is then performed three times evenly during the training process in the target domain.
Despite dropping in performance at the start of training after each state updating, mIoU gradually improves and eventually converges to a higher value than the previous round.
Figure \ref{fig:curve:b} shows the performance trend when the percentage of complex superpixels being excluded from learning gradually increases. As shown, the value mIoU constantly increases till reaching the level when 50\% of superpixels are filtered. In the remaining experiments, we fix the setting of RPT to involve 50\% of superpixels in regularization.

\begin{table}
      \centering
      \small
      \caption{\small Contribution of each design in RPT for domain adaptation of semantic segmentation on GTA5~$\to$~Cityscapes.}
      \begin{tabular}{l|c@{~}c@{~}c@{~}c@{~}c@{~}c|c} \hline
         \textbf{Method} & \textbf{ABN} & \textbf{ADV} & \textbf{PCR} & \textbf{CCR}  & \textbf{SLR} & \textbf{SU} & \textbf{mIoU} \\\hline
         FCN           &              &              &               &               &          &     & 32.3          \\
         +ABN         & $\surd$      &              &               &               &          &    & 39.1           \\
         FCN$_{adv}$ (+ADV)        & $\surd$      & $\surd$      &               &               &          &    &  47.2          \\ \hline
         +{PCR}         & $\surd$      & $\surd$      & $\surd$       &               &          &    & 49.0          \\
         +{CCR}        & $\surd$      & $\surd$      & $\surd$       & $\surd$       &          &     & 49.6          \\
         \textbf{RPT$^{1}$} (+{SLR})         & $\surd$      & $\surd$      & $\surd$       & $\surd$       & $\surd$  &    &  50.4          \\
         \textbf{RPT$^{3}$}           & $\surd$      & $\surd$      & $\surd$       & $\surd$       & $\surd$  &  $\surd$ &   52.6          \\\hline
      \end{tabular}
      \label{tab:contribution}
   \vspace{-0.15in}
\end{table}

\begin{table*}
   \centering
   \footnotesize
   \caption{\small Comparisons with the state-of-the-art unsupervised domain adaptation methods on GTA5~$\to$~Cityscapes adaptation. Please note that the baseline methods are divided into five groups: (1) representation-level domain adaptation by adversarial learning \cite{chen2018road,Du_2019_ICCV,pmlr-v80-hoffman18a,hong2018conditional,luo2019taking,sankaranarayanan2018learning,Tsai_2018_CVPR}; (2) appearance-level domain adaptation by image translation \cite{dundar2018domain,murez2018image}; (3) appearance-level + representation-level adaptation~\cite{chang2019all,wu2018dcan,Zhang_2018_CVPR}; (4) self-learning \cite{iqbal2019mlsl,lian2019constructing,zhang2018fully,zou2018unsupervised}; (5) others \cite{Chen_2019_ICCV,li2019bidirectional,saleh2018effective,zhang2017curriculum,zhu2018penalizing}.}
   \begin{tabular}{l@{~}|@{~}c@{~~}c@{~~}c@{~~}c@{~~}c@{~~}c@{~~}c@{~~}c@{~~}c@{~~}c@{~~}c@{~~}c@{~~}c@{~~}c@{~~}c@{~~}c@{~~}c@{~~}c@{~~}c@{~}|@{~}c@{~~}} \hline
      Method                                        & road          & sdwlk         & bldng         & wall          & fence         & pole          & light         & sign          & vgttn         & trrn          & sky           & person        & rider         & car           & truck         & bus           & train         & mcycl         & bcycl         & mIoU          \\ \hline
      FCNWild~\cite{BDDS_hoffman2016fcns}          & 70.4 & 32.4  & 62.1  & 14.9 & 5.4   & 10.9 & 14.2  & 2.7  & 79.2  & 21.3 & 64.6 & 44.1   & 4.2   & 70.4 & 8.0   & 7.3  & 0.0   & 3.5   & 0.0   & 27.1 \\
      Learning~\cite{sankaranarayanan2018learning} & 88.0 & 30.5  & 78.6  & 25.2 & 23.5  & 16.7 & 23.5  & 11.6 & 78.7  & 27.2 & 71.9 & 51.3   & 19.5  & 80.4 & 19.8  & 18.3 & 0.9   & 20.8  & 18.4  & 37.1 \\
      ROAD~\cite{chen2018road}                     & 76.3 & 36.1  & 69.6  & 28.6 & 22.4  & 28.6 & 29.3  & 14.8 & 82.3  & 35.3 & 72.9 & 54.4   & 17.8  & 78.9 & 27.7  & 30.3 & 4.0   & 24.9  & 12.6  & 39.4 \\
      CyCADA~\cite{pmlr-v80-hoffman18a}            & 79.1 & 33.1  & 77.9  & 23.4 & 17.3  & 32.1 & 33.3  & 31.8 & 81.5  & 26.7 & 69.0 & 62.8   & 14.7  & 74.5 & 20.9  & 25.6 & 6.9   & 18.8  & 20.4  & 39.5 \\
      AdaptSegNet~\cite{Tsai_2018_CVPR}            & 86.5 & 36.0  & 79.9  & 23.4 & 23.3  & 23.9 & 35.2  & 14.8 & 83.4  & 33.3 & 75.6 & 58.5   & 27.6  & 73.7 & 32.5  & 35.4 & 3.9   & 30.1  & 28.1  & 42.4 \\
      CLAN~\cite{luo2019taking}                    & 87.0 & 27.1  & 79.6  & 27.3 & 23.3  & 28.3 & 35.5  & 24.2 & 83.6  & 27.4 & 74.2 & 58.6   & 28.0  & 76.2 & 33.1  & 36.7 & 6.7   & 31.9  & 31.4  & 43.2 \\
      Conditional~\cite{hong2018conditional}       & 89.2 & 49.0  & 70.7  & 13.5 & 10.9  & 38.5 & 29.4  & 33.7 & 77.9  & 37.6 & 65.8 & \textbf{75.1}   & \textbf{32.4}  & 77.8 & \textbf{39.2}  & 45.2 & 0.0   & 25.5  & 35.4  & 44.5 \\
      SSF-DAN~\cite{Du_2019_ICCV}                  & 90.3 & 38.9  & 81.7  & 24.8 & 22.9  & 30.5 & 37.0  & 21.2 & 84.8  & 38.8 & 76.9 & 58.8   & 30.7  & 85.7 & 30.6  & 38.1 & 5.9   & 28.3  & 36.9  & 45.4 \\
      ADVENT~\cite{Vu_2019_CVPR}                   & 89.4 & 33.1  & 81.0  & 26.6 & 26.8  & 27.2 & 33.5  & 24.7 & 83.9  & 36.7 & 78.8 & 58.7   & 30.5  & 84.8 & 38.5  & 44.5 & 1.7   & 31.6  & 32.4  & 45.5 \\ \hline
      I2I Adapt~\cite{murez2018image}              & 85.8 & 37.5  & 80.2  & 23.3 & 16.1  & 23.0 & 14.5  & 9.8  & 79.2  & 36.5 & 76.4 & 53.4   & 7.4   & 82.8 & 19.1  & 15.7 & 2.8   & 13.4  & 1.7   & 35.7 \\
      Stylization~\cite{dundar2018domain}          & 86.9 & 44.5  & 84.7  & 38.8 & 26.6  & 32.1 & 42.3  & 22.5 & 84.7  & 30.9 & 85.9 & 67.0   & 28.1  & 85.7 & 38.3  & 31.8 & 21.5  & 31.3  & 24.6  & 47.8 \\ \hline
      DCAN~\cite{wu2018dcan}                       & 85.0 & 30.8  & 81.3  & 25.8 & 21.2  & 22.2 & 25.4  & 26.6 & 83.4  & 36.7 & 76.2 & 58.9   & 24.9  & 80.7 & 29.5  & 42.9 & 2.5   & 26.9  & 11.6  & 41.7 \\
      DISE~\cite{chang2019all}                     & 91.5 & 47.5  & 82.5  & 31.3 & 25.6  & 33.0 & 33.7  & 25.8 & 82.7  & 28.8 & 82.7 & 62.4   & 30.8  & 85.2 & 27.7  & 34.5 & 6.4   & 25.2  & 24.4  & 45.4 \\
      FCAN~\cite{Zhang_2018_CVPR}                  & 88.9 & 37.9  & 82.9  & 33.2 & 26.1  & \textbf{42.8} & 43.2  & 28.4 & 86.5  & 35.2 & 78.0 & 65.9   & 22.8  & 86.7 & 23.7  & 34.9 & 2.7   & 24.0  & 41.9  & 46.6 \\ \hline
      FCTN~\cite{zhang2018fully}                   & 72.2 & 28.4  & 74.9  & 18.3 & 10.8  & 24.0 & 25.3  & 17.9 & 80.1  & 36.7 & 61.1 & 44.7   & 0.0   & 74.5 & 8.9   & 1.5  & 0.0   & 0.0   & 0.0   & 30.5 \\
      CBST~\cite{zou2018unsupervised}              & 89.6 & \textbf{58.9}  & 78.5  & 33.0 & 22.3  & 41.4 & 48.2  & 39.2 & 83.6  & 24.3 & 65.4 & 49.3   & 20.2  & 83.3 & 39.0  & 48.6 & 12.5  & 20.3  & 35.3  & 47.0 \\
      PyCDA~\cite{lian2019constructing}            & \textbf{92.3} & 49.2  & 84.4  & 33.4 & 30.2  & 33.3 & 37.1  & 35.2 & 86.5  & 36.9 & 77.3 & 63.3   & 30.5  & 86.6 & 34.5  & 40.7 & 7.9   & 17.6  & 35.5  & 48.0 \\
      MLSL~\cite{iqbal2019mlsl}                    & 89.0 & 45.2  & 78.2  & 22.9 & 27.3  & 37.4 & 46.1  & 43.8 & 82.9  & 18.6 & 61.2 & 60.4   & 26.7  & 85.4 & 35.9  & 44.9 & \textbf{36.4}  & 37.2  & 49.3  & 49.0 \\ \hline
      Curriculum~\cite{zhang2017curriculum}        & 72.9 & 30    & 74.9  & 12.1 & 13.2  & 15.3 & 16.8  & 14.1 & 79.3  & 14.5 & 75.5 & 35.7   & 10    & 62.1 & 20.6  & 19   & 0     & 19.3  & 12    & 31.4 \\
      Penalizing~\cite{zhu2018penalizing}          & -    & -     & -     & -    & -     & -    & -     & -    & -     & -    & -    & -      & -     & -    & -     & -    & -     & -     & -     & 38.1 \\
      Effective~\cite{saleh2018effective}          & 79.8 & 29.3  & 77.8  & 24.2 & 21.6  & 6.9  & 23.5  & \textbf{44.2} & 80.5  & 38.0 & 76.2 & 52.7   & 22.2  & 83.0 & 32.3  & 41.3 & 27.0  & 19.3  & 27.7  & 42.5 \\
      MaxSquare~\cite{Chen_2019_ICCV}              & 89.3 & 40.5  & 81.2  & 29.0 & 20.4  & 25.6 & 34.4  & 19.0 & 83.6  & 34.4 & 76.5 & 59.2   & 27.4  & 83.8 & 38.4  & 43.6 & 7.1   & 32.2  & 32.5  & 45.2 \\
      Bidirectional~\cite{li2019bidirectional}     & 91.0 & 44.7  & 84.2  & 34.6 & 27.6  & 30.2 & 36.0  & 36.0 & 85.0  & \textbf{43.6} & 83.0 & 58.6   & 31.6  & 83.3 & 35.3  & \textbf{49.7} & 3.3   & 28.8  & 35.6  & 48.5 \\ \hline\hline
      \textbf{FCN$_{adv}$+RPT$^{1}$}                                  & 88.7 & 37.0  & 85.2  & 36.6 & 27.7  & 42.6 & 49.1  & 30.0 & 86.9  & 37.6 & 80.7 & 66.8   & 27.5  & 88.1 & 30.3  & 39.5 & 22.5  & 28.0  & 53.0  & 50.4 \\
      \textbf{FCN$_{adv}$+RPT$^{3}$}                                  & 89.2 & 43.3  & 86.1  & 39.5 & 29.9  & 40.2 & 49.6  & 33.1 & 87.4  & 38.5 & 86.0 & 64.4   & 25.1  & 88.5 & 36.6  & 45.8 & 23.9  & 36.5  & 56.8  & 52.6 \\
      \textbf{FCN$_{adv}$+RPT$^{3}$}+MS                               & 89.7 & 44.8  & \textbf{86.4}  & \textbf{44.2} & \textbf{30.6}  & 41.4 & \textbf{51.7}  & 33.0 & \textbf{87.8}  & 39.4 & \textbf{86.3} & 65.6   & 24.5  & \textbf{89.0} & 36.2  & 46.8 & 17.6 & \textbf{39.1} & \textbf{58.3} & \textbf{53.2} \\ \hline
   \end{tabular}
   \label{tab:GTA5}
   \vspace{-0.15in}
\end{table*}

\subsection{An Ablation Study}
\begin{figure}[!tb]
   \centering {\includegraphics[width=0.48\textwidth]{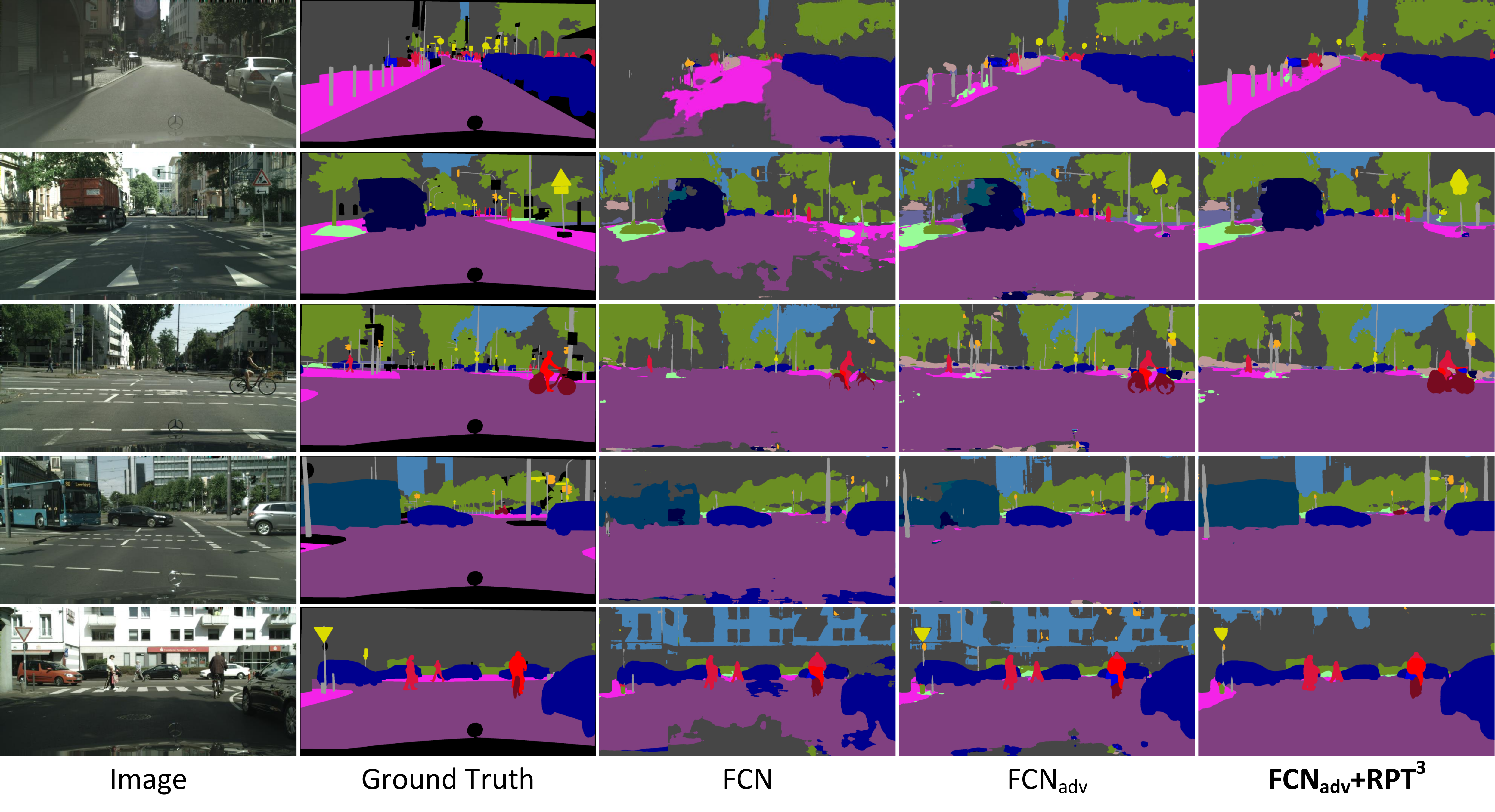}}
   \caption{\small Examples of semantic segmentation results on GTA5-Cityscapes adaptation. The original images, their ground truth and comparative results at different stages of FCN$_{adv}$+RPT$^{3}$ are given.}
   \label{fig:comparison}
   \vspace{-0.15in}
\end{figure}

Next, we conduct an ablation study to assess the performance impacts of different design components. We separately assess the three regularizations in RPT: patch-based consistency regularization (\textbf{PCR}), cluster-based consistency regularization (\textbf{CCR}) and spatial logic regularization (\textbf{SLR}). Table \ref{tab:contribution} details the contribution of each component towards the overall performance. FCN$_{adv}$, by considering adaptive batch normalization and adversarial learning (ABN+ADV), successfully boosts mIoU from 32.3\% to 47.2\%. The result indicates the importance of narrowing the domain gap between synthetic data and real images. The three regularizations in target domain introduce 1.8\%, 0.6\% and 0.8\% of improvement, respectively. Furthermore, by increasing the number of state updating during network optimization, additional 2.2\% of improvement is observed from RPT$^{1}$ to RPT$^{3}$. Figure \ref{fig:comparison} shows the gradual improvement on semantic segmentation of five images, when different design components are incrementally integrated.

\subsection{Comparisons with State-of-the-Art}
We compare with several state-of-the-art techniques for unsupervised domain adaptation on GTA5~$\to$~Cityscapes. Broadly, we can categorize the baseline methods into five categories: (1) representation-level domain adaptation by adversarial learning \cite{chen2018road,Du_2019_ICCV,pmlr-v80-hoffman18a,hong2018conditional,luo2019taking,sankaranarayanan2018learning,Tsai_2018_CVPR}; (2) appearance-level domain adaptation by image translation~\cite{dundar2018domain,murez2018image}; (3) appearance-level + representation-level adaptation \cite{chang2019all,wu2018dcan,Zhang_2018_CVPR}; (4) self-learning \cite{iqbal2019mlsl,lian2019constructing,zhang2018fully,zou2018unsupervised}; (5) others \cite{Chen_2019_ICCV,li2019bidirectional,saleh2018effective,zhang2017curriculum,zhu2018penalizing}. The performance comparisons on GTA5~$\to$~Cityscapes adaptation are summarized in Table~\ref{tab:GTA5}.
FCN$_{adv}$+RPT$^{3}$ achieves new state-of-the-art performance with mIoU of 52.6\%. Benefiting from the proposed regularizations, FCN$_{adv}$+RPT$^{3}$ outperforms SSF-DAN~\cite{Du_2019_ICCV} and ADVENT~\cite{Vu_2019_CVPR}, which also adopt a similar adversarial mechanism, by additional improvement of 7.2\% and 7.1\%, respectively. The performance is also better than the most recently proposed FCAN~\cite{Zhang_2018_CVPR} and Stylization~\cite{dundar2018domain}, which exploit a novel appearance transferring module that is not considered in RPT. Comparing to the best reported result to-date by MLSL~\cite{iqbal2019mlsl}, our proposed model still leads the performance by 3.6\%. By further integrating with the multi-scale (MS) scheme, i.e, FCN$_{adv}$+RPT$^{3}$+MS, the mIoU boosts to 53.2\% with 9 out of the 19 categories reach to-date the best reported performances.

To verify the generalization of RPT, we also test the performance on SYNTHIA~$\to$~Cityscapes using the same settings. Following previous works~\cite{iqbal2019mlsl,lian2019constructing,Vu_2019_CVPR,zou2018unsupervised}, the performances are reported in terms of mIoU@16 and mIoU@13 by not considering the different number of categories. The performance comparisons are summarized in Table \ref{tab:SYNTHIA}. Similarly, FCN$_{adv}$+RPT$^{3}$+MS achieves the best performance with mIoU@16 = 51.7\% and mIoU@13 = 59.5\%. The performances are better than PyCDA, which reports the best known results, by 5\% and 6.2\% respectively.
\begin{table*}[]
   \centering
   \footnotesize
   \caption{\small Comparisons with the state-of-the-art unsupervised domain adaptation methods on SYNTHIA~$\to$~Cityscapes transfer.}
   \begin{tabular}{l@{~}|@{~}c@{~~}c@{~~}c@{~~}c@{~~}c@{~~}c@{~~}c@{~~}c@{~~}c@{~~}c@{~~}c@{~~}c@{~~}c@{~~}c@{~~}c@{~~}c@{~}|@{~}c@{~~}c@{~}}\hline
                                                & road  & sdwlk & bldng & wall  & fence & pole  & light & sign  & vgttn & sky   & person & rider & car   & bus   & mcycl & bcycl & mIoU@16   & mIoU@13 \\ \hline
   Learning~\cite{sankaranarayanan2018learning} & 80.1          & 29.1          & 77.5          & 2.8           & 0.4          & 26.8          & 11.1          & 18.0          & 78.1          & 76.7          & 48.2          & 15.2          & 70.5          & 17.4          & 8.7           & 16.7          & 36.1          & -             \\
   ROAD~\cite{chen2018road}                     & 77.7          & 30.0          & 77.5          & 9.6           & 0.3          & 25.8          & 10.3          & 15.6          & 77.6          & 79.8          & 44.5          & 16.6          & 67.8          & 14.5          & 7.0           & 23.8          & 36.2          & -             \\
   AdaptSegNet~\cite{Tsai_2018_CVPR}            & 84.3          & 42.7          & 77.5          & -             & -            & -             & 4.7           & 7.0           & 77.9          & 82.5          & 54.3          & 21.0          & 72.3          & 32.2          & 18.9          & 32.3          & -             & 46.7          \\
   CLAN~\cite{luo2019taking}                    & 81.3          & 37.0          & 80.1          & -             & -            & -             & 16.1          & 13.7          & 78.2          & 81.5          & 53.4          & 21.2          & 73.0          & 32.9          & 22.6          & 30.7          & -             & 47.8          \\
   Conditional~\cite{hong2018conditional}       & 85.0          & 25.8          & 73.5          & 3.4           & \textbf{3.0} & 31.5          & 19.5          & 21.3          & 67.4          & 69.4          & 68.5          & 25.0          & 76.5          & 41.6          & 17.9          & 29.5          & 41.2          & -             \\
   SSF-DAN~\cite{Du_2019_ICCV}                  & 84.6          & 41.7          & 80.8          & -             & -            & -             & 11.5          & 14.7          & 80.8          & 85.3          & 57.5          & 21.6          & 82.0          & 36.0          & 19.3          & 34.5          & -             & 50.0          \\
   ADVENT~\cite{Vu_2019_CVPR}                   & 85.6          & 42.2          & 79.7          & 8.7           & 0.4          & 25.9          & 5.4           & 8.1           & 80.4          & 84.1          & 57.9          & 23.8          & 73.3          & 36.4          & 14.2          & 33.0          & 41.2          & 48.0          \\ \hline
   DCAN~\cite{wu2018dcan}                       & 82.8          & 36.4          & 75.7          & 5.1           & 0.1          & 25.8          & 8.0           & 18.7          & 74.7          & 76.9          & 51.1          & 15.9          & 77.7          & 24.8          & 4.1           & 37.3          & 38.4          & -             \\
   DISE~\cite{chang2019all}                     & \textbf{91.7} & \textbf{53.5} & 77.1          & 2.5           & 0.2          & 27.1          & 6.2           & 7.6           & 78.4          & 81.2          & 55.8          & 19.2          & 82.3          & 30.3          & 17.1          & 34.3          & 41.5          & -             \\\hline
   CBST~\cite{zou2018unsupervised}              & 53.6          & 23.7          & 75.0          & 12.5          & 0.3          & 36.4          & 23.5          & 26.3          & 84.8          & 74.7          & 67.2          & 17.5          & 84.5          & 28.4          & 15.2          & 55.8          & 42.5          & 48.4          \\
   PyCDA~\cite{lian2019constructing}            & 75.5          & 30.9          & 83.3          & 20.8          & 0.7          & 32.7          & 27.3          & \textbf{33.5} & 84.7          & 85.0          & 64.1          & 25.4          & 85.0          & 45.2          & 21.2          & 32.0          & 46.7          & 53.3          \\
   MLSL~\cite{iqbal2019mlsl}                    & 59.2          & 30.2          & 68.5          & \textbf{22.9} & 1.0          & 36.2          & 32.7          & 28.3          & \textbf{86.2} & 75.4          & \textbf{68.6} & 27.7          & 82.7          & 26.3          & 24.3          & 52.7          & 45.2          & 51.0          \\ \hline
   Curriculum~\cite{zhang2017curriculum}        & 57.4          & 23.1          & 74.7          & 0.5           & 0.6          & 14.0          & 5.3           & 4.3           & 77.8          & 73.7          & 45.0          & 11.0          & 44.8          & 21.2          & 1.9           & 20.3          & 29.7          & -             \\
   Penalizing~\cite{zhu2018penalizing}          & -             & -             & -             & -             & -            & -             & -             & -             & -             & -             & -             & -             & -             & -             & -             & -             & 34.2          & 40.3          \\
   MaxSquare~\cite{Chen_2019_ICCV}              & 82.9          & 40.7          & 80.3          & 10.2          & 0.8          & 25.8          & 12.8          & 18.2          & 82.5          & 82.2          & 53.1          & 18.0          & 79.0          & 31.4          & 10.4          & 35.6          & 41.4          & 48.2          \\
   Bidirectional~\cite{li2019bidirectional}     & 86.0          & 46.7          & 80.3          & -             & -            & -             & 14.1          & 11.6          & 79.2          & 81.3          & 54.1          & \textbf{27.9} & 73.7          & 42.2          & 25.7          & 45.3          & -             & 51.4          \\   \hline\hline
   \textbf{FCN$_{adv}$+RPT$^{1}$}                                  & 87.7          & 43.1          & 84.0          & 10.5          & 0.5          & \textbf{42.2} & \textbf{40.5} & 33.1          & 86.0          & 81.9          & 56.0          & 26.1          & 85.9          & 35.8          & 24.8          & 56.2          & 49.6          & 57.0        \\
   \textbf{FCN$_{adv}$+RPT$^{3}$}                                  & 88.9          & 46.5          & 84.5          & 15.1          & 0.5          & 38.5          & 39.5          & 30.1          & 85.9          & 85.8          & 59.8          & 26.1          & 88.1          & 46.8          & 27.7          & 56.1          & 51.2          & 58.9         \\
   \textbf{FCN$_{adv}$+RPT$^{3}$}+MS                               & 89.1          & 47.3          & \textbf{84.6} & 14.5          & 0.4          & 39.4          & 39.9          & 30.3          & 86.1          & \textbf{86.3} & 60.8          & 25.7          & \textbf{88.7} & \textbf{49.0} & \textbf{28.4} & \textbf{57.5} & \textbf{51.7} & \textbf{59.5}\\ \hline
   \end{tabular}
   \vspace{-0.15in}
   \label{tab:SYNTHIA}
\end{table*}
\subsection{Examples of Regularization}
\begin{figure}[!tb]
   \centering {\includegraphics[width=0.478\textwidth]{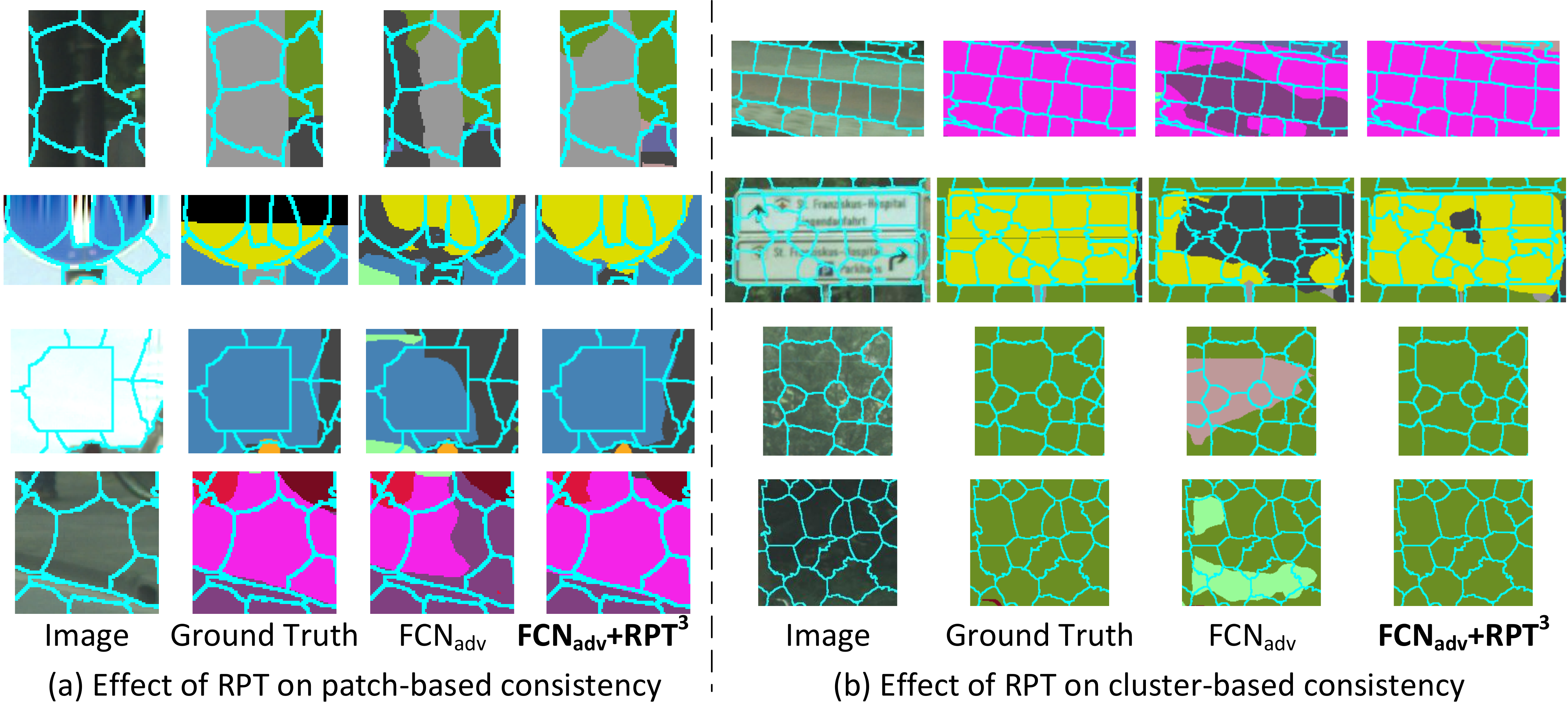}}
   \caption{\small Examples showing the effectiveness of patch-based consistency and cluster-based consistency in RPT.}
   \label{fig:case_loss}
   \vspace{-0.15in}
\end{figure}
\begin{figure}[!tb]
   \centering {\includegraphics[width=0.45\textwidth]{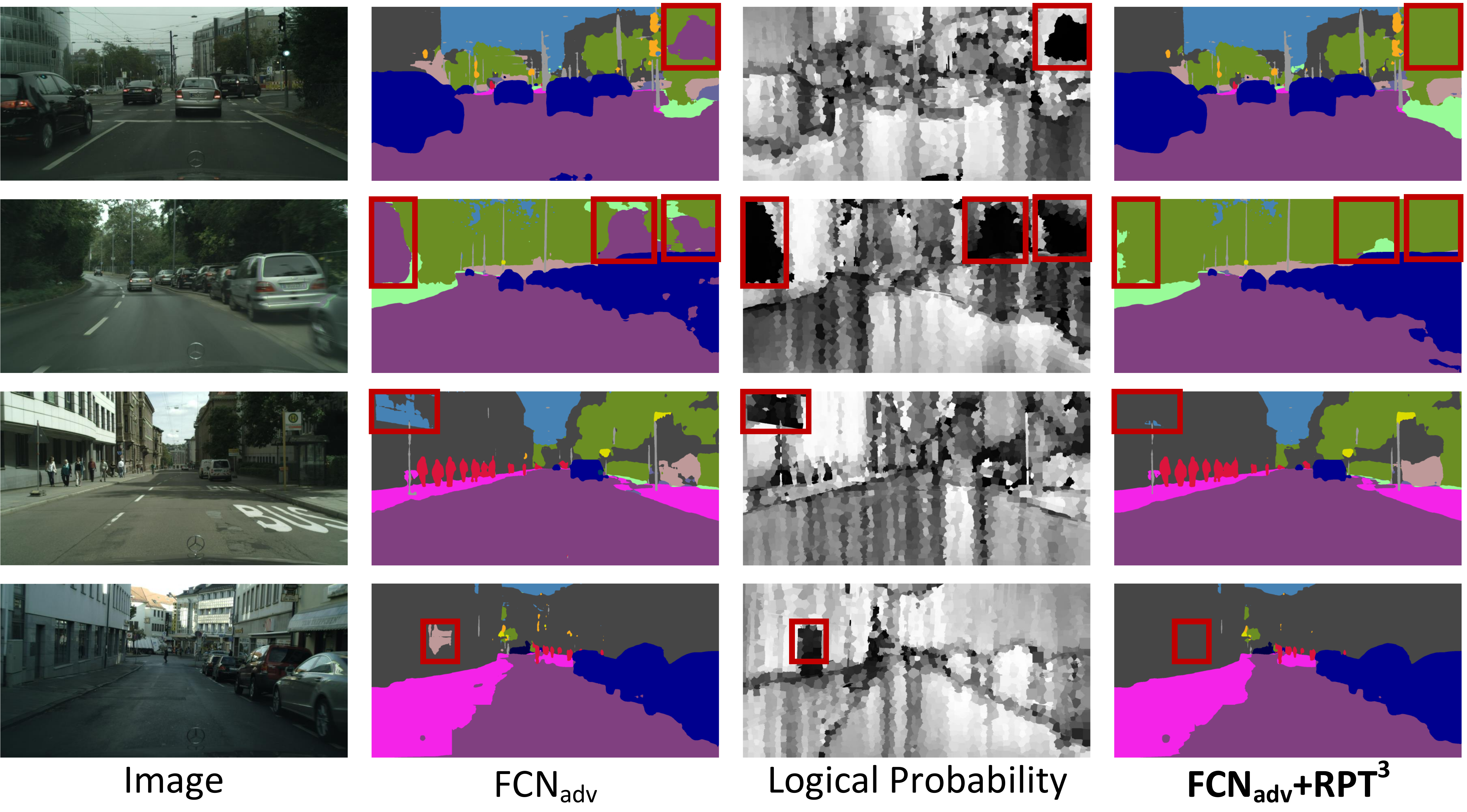}}
   \caption{\small The examples of punished patches by spatial logic.}
   \label{fig:case_loss_logic}
   \vspace{-0.15in}
\end{figure}

Figure~\ref{fig:case_loss} shows examples to demonstrate the effectiveness of patch-based and cluster-based consistency regularizations. Here, we crop some highlighted regions of input image, ground truth, prediction by FCN$_{adv}$ and prediction by FCN$_{adv}$+RPT$^{3}$, respectively.
On one hand, as shown in Figure~\ref{fig:case_loss}(a), patch-based consistency encourages the pixels to be predicted as the dominative category of the superpixel. On the other hand, cluster-based consistency is able to correct the predictions with the cue of visual similarity across superpixels as illustrated in Figure~\ref{fig:case_loss}(b).
These examples validate our motivation of enforcing label consistency within superpixel and cluster, where most semantic labels are correctly predicted in the target domain. Figure~\ref{fig:case_loss_logic} further visualizes the merit of modeling spatial context by spatial logic regularization.
Given the segmentation results from FCN$_{adv}$, our proposed LSTM encoder-decoder outputs the logical probability of assigning current semantic labels to each region. The darkness indicates that the region is predicted with low logical probability.
Better results are achieved by penalizing the illogical predictions, such as \textit{road} on the top of \textit{vegetation} (1$st$ row) or \textit{car} (2$nd$ row), \textit{sky} below \textit{building} (3$rd$ row), \textit{fence} above \textit{building} (4$th$ row).

\section{Conclusion}
We have presented Regularizer of Prediction Transfer (RPT) for unsupervised domain adaptation of semantic segmentation. RPT gives light to a novel research direction, by directly exploring the three intrinsic criteria of semantic segmentation to restrict the label prediction on the target domain. These criteria, when imposed as regularizers during training, are found to be effective in alleviating the problem of model overfitting.
The patch-based consistency attempts to unify the prediction inside each region by introducing its dominative category to the unconfident pixels. The cluster-based consistency further amends the prediction according to other visually similar regions which belong to the same cluster. In pursuit of suppressing illogical predictions, spatial logic is involved to regularize the spatial relation which is shared across domains.
Experiments conducted on the transfer from GTA5 to Cityscapes show that the injection of RPT can consistently improve the domain adaptation across different network architectures. More remarkably, the setting of FCN$_{adv}$+RPT$^{3}$ achieves new state-of-the-art performance. A similar conclusion is also drawn from the adaptation from SYNTHIA to Cityscapes, which demonstrates the generalization ability of RPT.

{\small
\bibliographystyle{ieee_fullname}
\bibliography{egbib}
}

\end{document}